\documentclass[10pt,twocolumn,letterpaper]{article}

\usepackage{iccv}              

\usepackage[accsupp]{axessibility}  

\newcommand{\yf}[1]{}

\definecolor{iccvblue}{rgb}{0.21,0.49,0.74}
\usepackage[pagebackref,breaklinks,colorlinks,allcolors=iccvblue]{hyperref}

\def\paperID{13618} 
\def\confName{ICCV}
\def\confYear{2025}

\title{Capturing head avatar with hand contacts from a monocular video}

\author{
Haonan He$^{1}$ 
\quad Yufeng Zheng$^{3,4}$
\quad Jie Song$^{1,2}$ \\
 $^1$The Hong Kong University of Science and Technology (Guangzhou) \\
 $^2$The Hong Kong University of Science and Technology \\
 $^3$ETH Z{\"u}rich, Switzerland
 \quad $^4$Max Planck Institute for Intelligent Systems, T{\"u}bingen, Germany \\
}

\begin{document}

\maketitle


\begin{abstract}
Photorealistic 3D head avatars are vital for telepresence, gaming, and VR. However, most methods focus solely on facial regions, ignoring natural hand-face interactions, such as a hand resting on the chin or fingers gently touching the cheek, which convey cognitive states like pondering. In this work, we present a novel framework that jointly learns detailed head avatars and the non-rigid deformations induced by hand-face interactions.
There are two principal challenges in this task. First, naively tracking hand and face separately fails to capture their relative poses. To overcome this, we propose to combine depth order loss with contact regularization during pose tracking, ensuring correct spatial relationships between the face and hand. Second, no publicly available priors exist for hand-induced deformations, making them non-trivial to learn from monocular videos. To address this, we learn a PCA basis specific to hand-induced facial deformations from a face-hand interaction dataset. This reduces the problem to estimating a compact set of PCA parameters rather than a full spatial deformation field. Furthermore, inspired by physics-based simulation, we incorporate a contact loss that provides additional supervision, significantly reducing interpenetration artifacts and enhancing the physical plausibility of the results.
We evaluate our approach on RGB(D) videos captured by an iPhone. Additionally, to better evaluate the reconstructed geometry, we construct a synthetic dataset of avatars with various types of hand interactions. We show that our method can capture better appearance and more accurate deforming geometry of the face than SOTA surface reconstruction methods. 
\end{abstract}
\begin{figure}[t]
    \centerline{\includegraphics[width=\linewidth]{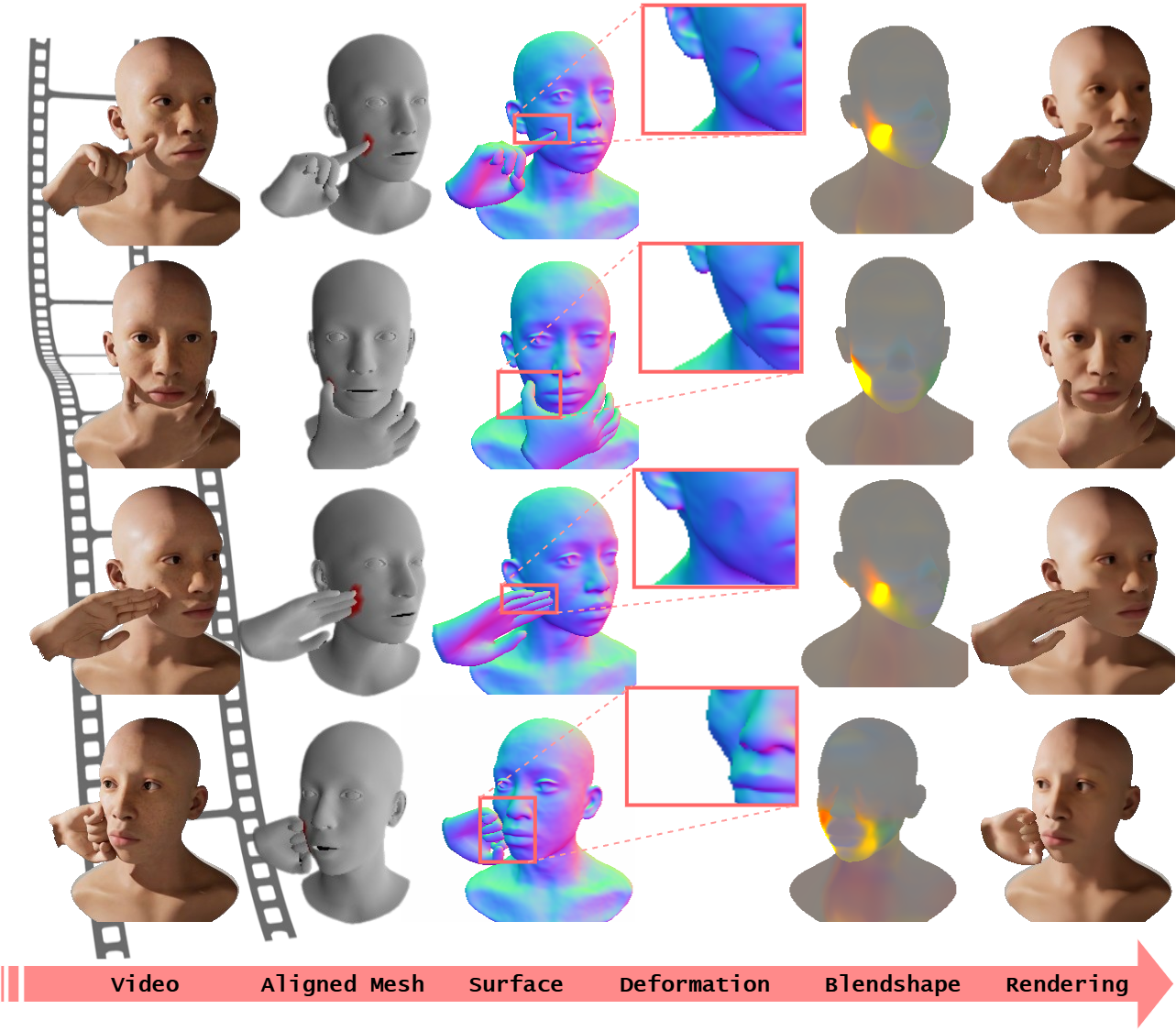}}
    \caption{
        Given an RGB video capturing hand-face interactions, our method automatically aligns tracked hand and face meshes, reconstructs high-fidelity 3D surfaces, and renders photorealistic textures. We further model contact-induced non-rigid deformations through learned blendshape fields guided by a non-rigid deformation PCA prior derived from hand-face interaction data.}
    \label{fig:teaser}
\end{figure}%
\vspace{-10pt}    
\section{Introduction}
\label{sec:intro}

How often do you touch your face throughout the day? Research indicates that people frequently touch their faces – averaging about 50 touches per hour~\cite{Rahman2020How}. This high frequency underscores the importance of hand-face interactions as subtle yet critical cues in everyday nonverbal communication, although they often occur unconsciously.

In recent years, the reconstruction of 3D head avatars from video data has received significant attention, driven by applications in telepresence, gaming, and virtual reality. However, most existing methods concentrate exclusively on facial or head reconstruction~\cite{Zheng2023pointavatar,bharadwaj2023flare,Duan2023bakedavatar,chen2024monogaussianavatar,xu2023gaussianheadavatar,qian2024gaussianavatars}, largely overlooking the dynamic interplay between the hand and face. This omission is critical, as hand-face interactions provide essential context for interpreting human behavior.

An early approach for modeling face-hand interactions~\cite{Shimada2023Decaf} attempts to predict hand-induced facial deformations on top of a 3D morphable model (3DMM). While innovative, its results lack the person-specific geometric details and realistic textures necessary for truly lifelike avatars. A more recent method, NePHIM~\cite{wagner2024nephim}, enhances fidelity by modeling personalized geometry; However, none have simultaneously produced head avatars with detailed geometry, high-quality texture, and physically plausible non-rigid deformations induced by hand contacts. In contrast, our method achieves all these objectives using only a monocular iPhone video.
In the following, we outline the two major challenges in this task and describe our solutions.

First, robust reconstruction of both the head and hand from a monocular video requires precise joint tracking of their poses. Conventional pipelines typically rely on separate estimations (e.g., DECA~\cite{feng2021learninganimatabledetailed3d} for head pose and expression and HaMeR~\cite{pavlakos2024hamer} for hand pose), but such independent tracking fails to capture the spatial relationship and contact dynamics between the face and hand. To address this, we incorporate depth information—
from off-the-shelf depth estimators—by applying a depth order loss that ensures nearby face and hand pixels are correctly ordered in depth. In addition, we introduce a contact regularization term that encourages plausible interactions when face and hand vertices are in close proximity. The depth order loss and contact regularization jointly ensure the correct relative positioning of the face and hand in the video.

The second challenge lies in modeling the non-rigid deformations that occur during hand-face interactions. Unlike expression-driven deformations—which can rely on established 3DMM priors—hand-induced facial deformations lack such guidance. We tackle this by first constraining the deformation space: we construct a PCA basis for hand-induced deformations using captured interaction data. This reduces the problem to estimating a compact set of PCA parameters rather than a full spatial deformation field. Moreover, we note that solely relying on RGB and mask losses is insufficient to learn accurate and plausible facial deformations. Inspired by physics-based simulations, we introduce a contact loss that mitigates face-hand interpenetration, thereby enhancing the physical plausibility of the reconstructed deformations.

We evaluate our approach on RGB(D) videos captured by an iPhone and further validate our reconstructed geometry using a synthetic dataset of avatars with varied hand interactions. We only use the RGB channels when evaluating our methods on captured real videos. Our experiments demonstrate that our method not only enhances the visual realism of the head avatars but also more accurately captures the dynamic interplay between the hand and face, outperforming state-of-the-art reconstruction methods.

In summary, our contributions are as follows:
\begin{enumerate}
    \item We propose a novel framework that jointly reconstructs detailed 3D head avatars with realistic textures and person-specific geometry, while capturing physically plausible non-rigid deformations induced by hand-face interactions --- all from a monocular iPhone video.
    \item We introduce a joint tracking strategy that leverages a depth order loss and contact regularization to accurately capture the spatial relationships and dynamic contacts between the face and hand.
    \item We constrain the optimization of non-rigid facial deformations by constructing a PCA basis for hand-induced facial deformations, reducing the problem to estimating a compact set of PCA parameters, and further enforce physical plausibility with a physics-inspired contact loss.
    \item Extensive evaluations on both real RGB(D) videos and a synthetic dataset demonstrate that our approach outperforms state-of-the-art methods in terms of appearance fidelity and geometric accuracy.
\end{enumerate}

\section{Related Work}
\subsection{Monocular Dynamic Surface Reconstruction}
Reconstructing dynamic surfaces from monocular RGB-D videos is a highly under-constrained problem.  
Early methods, such as DynamicFusion~\cite{newcombe2015dynamicfusion} and KinectFusion~\cite{izadi2011kinectfusion}, estimate a template-free 6D motion field to warp live frames into a TSDF surface. Subsequent works address key limitations, including handling topological changes~\cite{slavcheva2017killingfusion,slavcheva2018sobolevfusion}, improving tracking for fast and complex motions~\cite{bozic2020deepdeform,bozic2020neural}, and mitigating occlusions~\cite{lin2022occlusionfusion}. NDR~\cite{cai2022neural} introduces an invertible bijective mapping between the observation space and canonical space for more robust motion tracking, while MorpheuS~\cite{wang2024morpheus} utilizes a diffusion prior to achieve full 360° surface reconstruction.

Unlike these methods, which focus on general dynamic surface tracking, our approach explicitly models hand-face interactions. By leveraging priors from head and hand 3DMMs, incorporating contact constraints, and enforcing deformation priors, our method achieves physically plausible reconstructions of the human face and hand.


\subsection{Hand-Face Interaction}
Few works have focused on modeling hand-face interactions on tracked meshes. DECAF~\cite{Shimada2023Decaf} was the first to reconstruct 3D hand-face interactions from images. Using a dataset of multi-view videos, they track both the face and hand while reconstructing coarse facial geometry through physics-based simulation. Additionally, they propose an end-to-end network to predict contact points and deformations.
DICE~\cite{wu2024dice} improves accuracy by incorporating additional training on in-the-wild images and leveraging a pre-trained depth estimator. NePHIM~\cite{wagner2024nephim} further enhances realism by utilizing personalized head templates and modeling skin pulling effects.

These methods primarily focus on predicting contacts and deformations from a single image, making them suboptimal for video-based hand-face tracking and reconstruction. In contrast, our work reconstructs photorealistic avatars with smooth, physically plausible hand-face interactions from video sequences.

\begin{figure*}[t]
    \centerline{\includegraphics[width=1\linewidth]{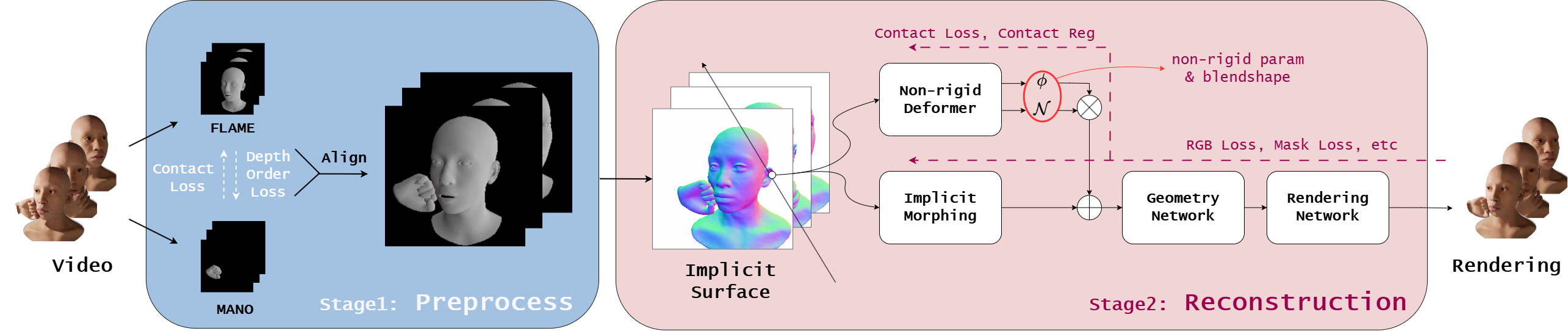}}
    \caption{
        \textbf{Pipeline of our method} Our framework operates through two stages: \textbf{Preprocessing} aligns separately tracked FLAME (face) and MANO (hand) meshes into a unified coordinate system via joint optimization of depth ordering loss and contact loss. \textbf{Reconstruction} learns neural deformation fields for the head avatar, with a contact-specific non-rigid deformation network. This specialized component, regularized by contact losses, explicitly models facial surface deformation induced by hand-face interaction.}
    \label{fig:pipeline}
\end{figure*}%

\section{Method}
Our framework consists of two core stages: preprocessing and reconstruction. During preprocessing (Sec.~\ref{sec:preprocessing}), we track hand and face meshes within a unified coordinate system and refine their relative positions using depth order loss and contact regularization. In the reconstruction stage (Sec.~\ref{sec:avatar_optimization}), we learn hand and face avatars with physically plausible non-rigid deformations from monocular RGB 
video. To regularize facial deformations, we leverage a PCA basis from a hand-face interaction dataset and enforce physically plausible hand-face contact dynamics via a contact loss.

\subsection{Hand-Face Mesh Alignment}
\label{sec:preprocessing}
We begin by estimating 3DMM parameters for the hand (MANO~\cite{romero2022embodied}) and head (FLAME~\cite{li2017learning}) in each video frame using DECA~\cite{feng2021learninganimatabledetailed3d} and HaMeR~\cite{pavlakos2024hamer}. Given an estimated perspective camera matrix, we refine the scale, shift, and pose parameters of both models by minimizing a 2D landmark loss~\cite{bulat2017far}. To track correct relative positions of the hand and the face, we introduce a depth order loss and a contact regularization term.
Specifically, we randomly sample pixels within the hand and face regions and query their respective depth values from both the rendered depth map of the tracked 3DMMs $\left(\hat{p_{h}}, \hat{p_{f}}\right)$ and the estimated depth map $\left(p_{h}, p_{f}\right)$ obtained from a pretrained depth estimator~\cite{yang2024depthv2}. The depth order loss $\mathcal{L}_{\text{order}}$ enforces correct relative depth ordering:

\begin{equation}
 \mathcal{L}_{\text{order}} = \max\left(0,\, - \text{sign}(p_{h} - p_{f}) \cdot (\hat{p_{h}} - \hat{p_{f}})\right).
\end{equation}

Additionally, we introduce a contact regularization term to encourage fingertip vertices to maintain contact with the closest facial vertices:
\begin{equation}
 \mathcal{L}_{\text{contact}} = \frac{1}{N \cdot K} \sum_{i=1}^{N} \sum_{k=1}^{K} \left\| \mathbf{v}_i^h - \mathbf{u}_{i_k}^f \right\|^2,
\end{equation}
where $N$ represents the set of fingertip vertices, and $K$ consists of facial vertices in contact-prone areas such as the cheeks, chin, and nose.



By jointly optimizing these losses along with projected landmark loss and a temporal smoothness regularization, we achieve accurate alignment and robust tracking of the relative positions of the hand and face meshes.

\subsection{Neural Implicit Avatar}
\label{sec:avatar_optimization}
\textbf{Face Avatar.} We represent face avatars using deformable neural implicit fields, modeled by three networks: a canonical geometry network, a canonical rendering network, and a deformation network. Below, we outline the rendering process step by step.

Given a pixel and camera projection matrix, we follow IDR~\cite{yariv2020multiview} to sample points $x_d$ along a ray. To map $x_d$ to the canonical space, we estimate FLAME~\cite{li2017learning} blendshapes for each deformed point and remove the expression-induced deformation to obtain the corresponding canonical point $x_c$. Specifically, our deformation network $f_{\sigma_d}$ predicts additive expression blendshape vectors $\mathcal{E} \in \mathbb{R}^{n_e \times 3}$, pose correctives $\mathcal{P} \in \mathbb{R}^{n_j \times 9 \times 3}$, and linear blend skinning weights $\mathcal{W} \in \mathbb{R}^{n_j}$:

\begin{equation}
f_{\sigma_d}(x_d, \boldsymbol{\theta}, \boldsymbol{\psi}) : \mathbb{R}^3\times \mathbb{R}^{15}\times \mathbb{R}^{50} \to \mathcal{E},\mathcal{P}, \mathcal{W}.
\end{equation}

The canonical correspondence $x_c$ is then computed as:

\begin{equation}
x_c = LBS^{-1}\left(x_d, J(\boldsymbol{\psi}), \boldsymbol{\theta}, \mathcal{W}\right) - B_E(\boldsymbol{\psi}; \mathcal{E}) - B_P(\boldsymbol{\theta}; \mathcal{P}),
\label{eq:inverse_lbs}
\end{equation}

where $\boldsymbol{\psi}$ and $\boldsymbol{\theta}$ are the expression and pose parameters, and $J$ is the FLAME joint regressor. $B_E(\cdot)$ and $B_P(\cdot)$ compute the expression and pose offsets using predicted blendshapes and pose correctives $\mathcal{E}$ and $\mathcal{P}$, and $LBS^{-1}(\cdot)$ undo joint rotations with predicted skinning weights $\mathcal{W}$.  

Next, the canonical geometry network $f_{\sigma_g}$ predicts the face occupancy value:

\begin{equation}
f_{\sigma_g}(x_c): \mathbb{R}^3 \to occ_f.
\end{equation}

We iteratively locate the ray-surface intersection point where ${occ_f} = 0.5$. We denote the canonical surface intersection point as $x_c^*$ and its deformed counterpart as $x_d^*$ from now on. 

After identifying the ray-surface intersections, we compute the normal direction $n_d^f$ of the deformed surface and use the rendering MLP $f_{\sigma_r}$ to obtain the final RGB value:

\begin{equation}
f_{\sigma_r}(x_c^*, n_d^f, \boldsymbol{\theta}, \boldsymbol{\psi}) : \mathbb{R}^3 \times \mathbb{R}^3 \times \mathbb{R}^{15}\times \mathbb{R}^{50} \to c_f.
\end{equation}

\textbf{Hand Avatar.} Since hand geometry is similar across subjects, we use the tracked MANO mesh to represent the dynamic hand geometry. For convenient joint rendering of the face and the hand, we convert the MANO mesh to an occupancy field: 
\begin{equation}
f_{h}(x): \mathbb{R}^3 \to {occ_h}.
\end{equation}

Similar to the face texture network, we represent the texture of hand mesh using a texture MLP $f_{\sigma_t}$ to map $x_c$ and corresponding normal values of the surface point $n_d^h$ to RGB colors $c_h$:
\begin{equation}
f_{\sigma_t}(x_c, n_d^h): \mathbb{R}^3 \times \mathbb{R}^3 \to {c_h},
\end{equation}
where the normal values $n_d^h$ are sampled from the MANO mesh by interpolating vertex normals of the nearest face using barycentric weights.

\subsection{Contact-Induced Non-Rigid Deformation}
To model contact-induced facial deformations, we introduce an additional set of blendshapes for the face and jointly optimize both the blendshapes and contact parameters during training.

\textbf{Non-Rigid Deformation Network.} We use a non-rigid deformation network $f_{\sigma_n}$ to predict contact-related blendshapes $\mathcal{N}$:

\begin{equation}
f_{\sigma_n}(x_c, l): \mathbb{R}^3 \times \mathbb{R}^{30} \to \mathcal{N},
\end{equation}

where $l$ is a per-frame optimizable latent code. Additionally, we estimate per-frame contact parameters $\boldsymbol{\phi} \in \mathbb{R}^{n_k}$, which scale the contact-related blendshapes $\mathcal{N} \in \mathbb{R}^{n_k} \times 3$ to obtain the contact-induced deformations. In practice, these parameters are also predicted from the latent code:

\begin{equation}
f_{\sigma_n}(l): \mathbb{R}^{30} \to \boldsymbol{\phi}.
\end{equation}

The canonical points in Eq.~\ref{eq:inverse_lbs} are then updated as:

\begin{equation}
  \begin{split}
    x_c ={} & LBS^{-1}\left(x_d, J(\boldsymbol{\psi}), \boldsymbol{\theta}, \mathcal{W}\right) \\
            & - B_E(\boldsymbol{\psi}; \mathcal{E}) - B_P(\boldsymbol{\theta}; \mathcal{P}) - B_N(\boldsymbol{\phi}; \mathcal{N}),
  \end{split}
  \label{eq:canonical_point}
\end{equation}
where $B_N(\cdot)$ computes additive offsets from contact-related blendshapes $\mathcal{N}$ and contact parameters $\boldsymbol{\phi}$.

\textbf{Non-Rigid Deformation PCA Prior.}
Since both the non-rigid blendshapes and contact parameters are unknown, the problem is highly under-constrained. To regularize optimization, we learn a PCA basis from a hand-face interaction dataset~\cite{Shimada2023Decaf}. Specifically, we extract per-frame non-rigid 3D displacements of FLAME vertices and construct a vertex deformation matrix. We perform PCA decomposition on this matrix, and retain the top $n_k$ components as our non-rigid basis. We then supervise the non-rigid blendshapes $\mathcal{N}$ using this prior, constraining optimization to a compact set of PCA parameters while promoting natural facial deformations caused by hand-face interactions.

\textbf{Contact Loss.} 
To prevent interpenetration and improve the physical plausibility of hand-face interactions, we introduce a contact loss $\mathcal{L}_{\text{contact}}$. Specifically, we sample points $x_d^i \in M_h$ on the hand surface and enforce that the face geometry does not occupy these regions:

\begin{equation*}
\mathcal{L}_{\text{contact}} = \frac{1}{|M_h|} \sum_{i \in {M_h}} \max\left(0, -f_{\sigma_g}(x_c^i)\right),
\end{equation*}
where $x_c^i$ is the canonical correspondence of sampled hand surface points $x_d^i$ (see Eq.~\ref{eq:canonical_point}).

Additionally, we introduce a regularization term to minimize non-rigid deformation in non-penetration regions:

\begin{equation*}
\mathcal{L}_{\text{reg}} = \frac{1}{|M_f|} \sum_{i \in M_f\setminus M} ||B_N(\boldsymbol{\phi}_i; \mathcal{N}_{i})||_2,
\end{equation*}
where $M_f$ consists of points randomly sampled around the deformed FLAME surface, and $M$ denotes points inside both face and hand geometry, where interpenetration exists. 
The contact and regularization losses only optimize the non-rigid deformation network and contact parameters, to avoid undesired gradient updates to the head geometry and expression-related deformations.

\subsection{Training Objectives}
Our method is supervised by multiple loss terms. The primary RGB loss~\eqref{eq:rgb} enforces photometric consistency by minimizing the $L_2$ distance between rendered colors $f_{\sigma_r}(x_c^*)$ and ground-truth pixel values $\mathbf{C}$ across foreground pixels: 
\begin{equation}
    \begin{split}
        \mathcal{L}_{\text{RGB}} ={} & \frac{1}{|P|} \sum_{i \in P^f} \|\mathbf{C}_i - f_{\sigma_r}(x_c^*)\|_2^2 \\
        & + \frac{1}{|P|} \sum_{i \in P^h} \|\mathbf{C}_i - f_{\sigma_t}(x_c^*)\|_2^,
        \label{eq:rgb}
    \end{split}
\end{equation}
where $P$ denotes all training pixels, $P^{f}$ is the set of rays in the intersection of the estimated face mask $O_f^i$ and rendered face occupancy, and similarly, $P^{h}$ denotes the intersection region for the hand.  
To supervise the face geometry, we also employ a mask loss~\eqref{eq:mask} that applies cross-entropy (CE) supervision on the predicted occupancy values $f_{\sigma_g}(x_c)$. This is guided by a pseudo ground-truth head mask $O_f^i$, while excluding pixels within the hand mask $O_h^i$ to avoid wrong supervision in occluded regions:

\begin{equation}
    \mathcal{L}_{\text{M}} = \frac{1}{|P|} \sum_{i \in P \setminus (P_f, O_h^i)} \text{CE}(O_f^i, f_{\sigma_g}(x_c^i)). 
    \label{eq:mask}
\end{equation}

To incorporate facial prior knowledge, we introduce a FLAME loss~\eqref{eq:lbs} that aligns predicted blendshapes and skinning weights ($\mathcal{E}_i$, $\mathcal{P}_i$, $\mathcal{W}_i$) with pseudo ground-truth values from the nearest FLAME vertices. Additionally, we constrain the non-rigid blendshape vectors $\mathcal{N}$ using the PCA basis $\mathcal{N}^{\text{GT}}$ derived from a hand-face interaction dataset:

\begin{equation}
    \begin{split}
        \mathcal{L}_{\text{lbs}} ={} & \frac{1}{|P|} \sum_{i \in P_f} \left[ \lambda_{e}\|\mathcal{E}_i - \mathcal{E}_i^{\text{GT}}\|_2^2 + \lambda_{p}\|\mathcal{P}_i - \mathcal{P}_i^{\text{GT}}\|_2^2 \right. \\
        & \left. + \lambda_{w}\|\mathcal{W}_i - \mathcal{W}_i^{\text{GT}}\|_2^2 + \lambda_{n}\|\mathcal{N}_i - \mathcal{N}_i^{\text{GT}}\|_2^2 \right],
        \label{eq:lbs}
    \end{split}
\end{equation}

with weighting factors $\lambda_e = 1000$, $\lambda_p = 1000$, $\lambda_w = 0.1$, and $\lambda_n = 10000$.

The final objective~\eqref{eq:total} combines all loss terms:

\begin{equation}
    \mathcal{L}_{\text{total}} = \mathcal{L}_{\text{RGB}} + \lambda_M\mathcal{L}_{\text{M}} + \lambda_{\text{lbs}}\mathcal{L}_{\text{lbs}} + \lambda_{\text{contact}}\mathcal{L}_{\text{contact}} + \lambda_{\text{reg}}\mathcal{L}_{\text{reg}},
    \label{eq:total}
\end{equation}

where $\lambda_M = 2$, $\lambda_{\text{lbs}} = 1$, $\lambda_{\text{contact}} = 1000$, and $\lambda_{\text{reg}}=10$ balance the contributions of each term.

\begin{figure*}[t]
    \centerline{\includegraphics[width=0.8\linewidth]{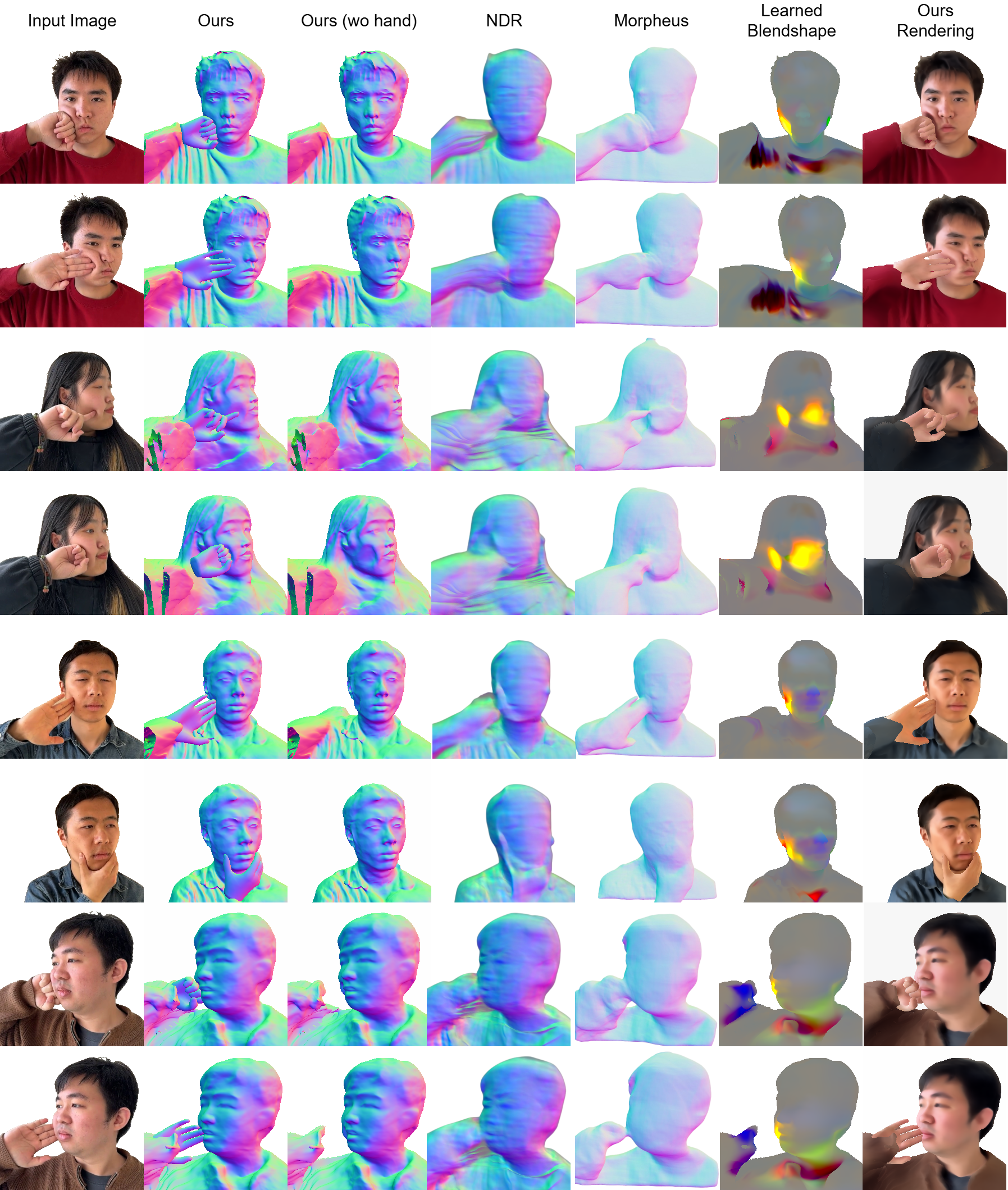}}
    \caption{
        \textbf{Qualitative Comparison with {SOTA}\xspace on Captured Videos} Reconstruction results from captured video sequences comparing our method with {SOTA}\xspace baselines. Our method trains exclusively on RGB video input, whereas NDR and Morpheus require LiDAR-derived depth maps. Qualitative comparisons demonstrate our approach achieves superior hand and face reconstruction fidelity while faithfully recovering physically plausible facial deformations from hand interactions. The final columns visualize our learned blendshape fields alongside photorealistic rendering outputs.}
    \label{fig:qualitative_comparison_real}
\end{figure*}%

\section{Experiments}

In this section, we compare our method with NDR~\cite{cai2022neural} and Morpheus~\cite{wang2024morpheus} on both real-world captured videos and our newly introduced synthetic dataset. Our results demonstrate the superior effectiveness of the proposed approach in accurately modeling both the observed surfaces and the occluded hand-face contact regions.

\subsection{Dataset}


\noindent\textbf{Synthetic Dataset} We introduce a synthetic dataset comprising 3 subjects performing 4 hand-face interaction sequences. Each subject is constructed using Unreal Engine 5's MetaHuman Creator plugin with photorealistic textures. Facial expressions and head poses are captured via iPhone Face ID to drive the facial animation system, while hand interaction sequences are manually designed to reflect natural contact patterns.

Non-rigid facial deformations resulting from hand contact are simulated through Position-Based Dynamics (PBD) implemented through Geometry Nodes in Blender. The dataset provides comprehensive multi-modal data including high-resolution rendered video sequences, segmentation masks, depth maps, surface normal maps, and ground-truth mesh tracking for both facial and hand components. Quantitative evaluation across multiple metrics demonstrates that our method can reconstruct more accurate facial geometry and deformation than state-of-the-art surface reconstruction methods.

\noindent\textbf{Real-video Dataset} 
We evaluate our method on four real-world video sequences capturing distinct hand-face interaction scenarios. All data was captured using the LiDAR sensor on an iPhone 15 Pro, with hand and face masks generated through off-the-shelf video segmentation methods \cite{cheng2024putting}. Each recording contains approximately 1,000 frames featuring a single subject performing four interaction tasks.
For 3D reconstruction, we estimate FLAME parameters using DECA \cite{feng2021learninganimatabledetailed3d} and MANO parameters through HaMer \cite{pavlakos2024hamer}. Facial keypoints were detected using \cite{bulat2017far}, while hand keypoints were extracted with Sapiens \cite{khirodkar2024sapiens}. To address Sapiens' limitations in detecting occluded thumb regions, we supplement these measurements with projected MANO mesh landmarks in these cases.


\subsection{Comparison on Real Videos}

We present qualitative comparisons between our method, NDR, and Morpheus on real-world captured videos in Fig. \ref{fig:qualitative_comparison_real}. Our approach successfully aligns hand and face meshes within a unified coordinate system while reconstructing high-quality avatars without requiring LiDAR depth maps. Notably, our results are derived solely from RGB video inputs, while NDR and Morpheus require RGB-D sensor data.

Our method produces more refined geometric details in both hand and facial surfaces. While NDR and Morpheus leverage depth information, they still fail to reconstruct accurate surface topologies due to inherent limitations in handling extensive motion variations. These comparative methods prove particularly underconstrained when processing dynamic sequences containing significant head rotations and diverse hand articulations.

A critical advantage of our technique lies in modeling occluded contact regions between hands and faces. As demonstrated in Fig.~\ref{fig:qualitative_comparison_real}, facial surfaces adaptively deform to create contoured indentations matching hand geometry. This contact-aware deformation propagates coherently through adjacent visible surfaces, achieving physically plausible deformations through joint optimization of visual constraints and geometric priors. The learned non-rigid deformation blendshape fields indicate where the contact happens and the shape of facial deformation. 


\begin{figure*}[t]
    \centerline{\includegraphics[width=0.8\linewidth]{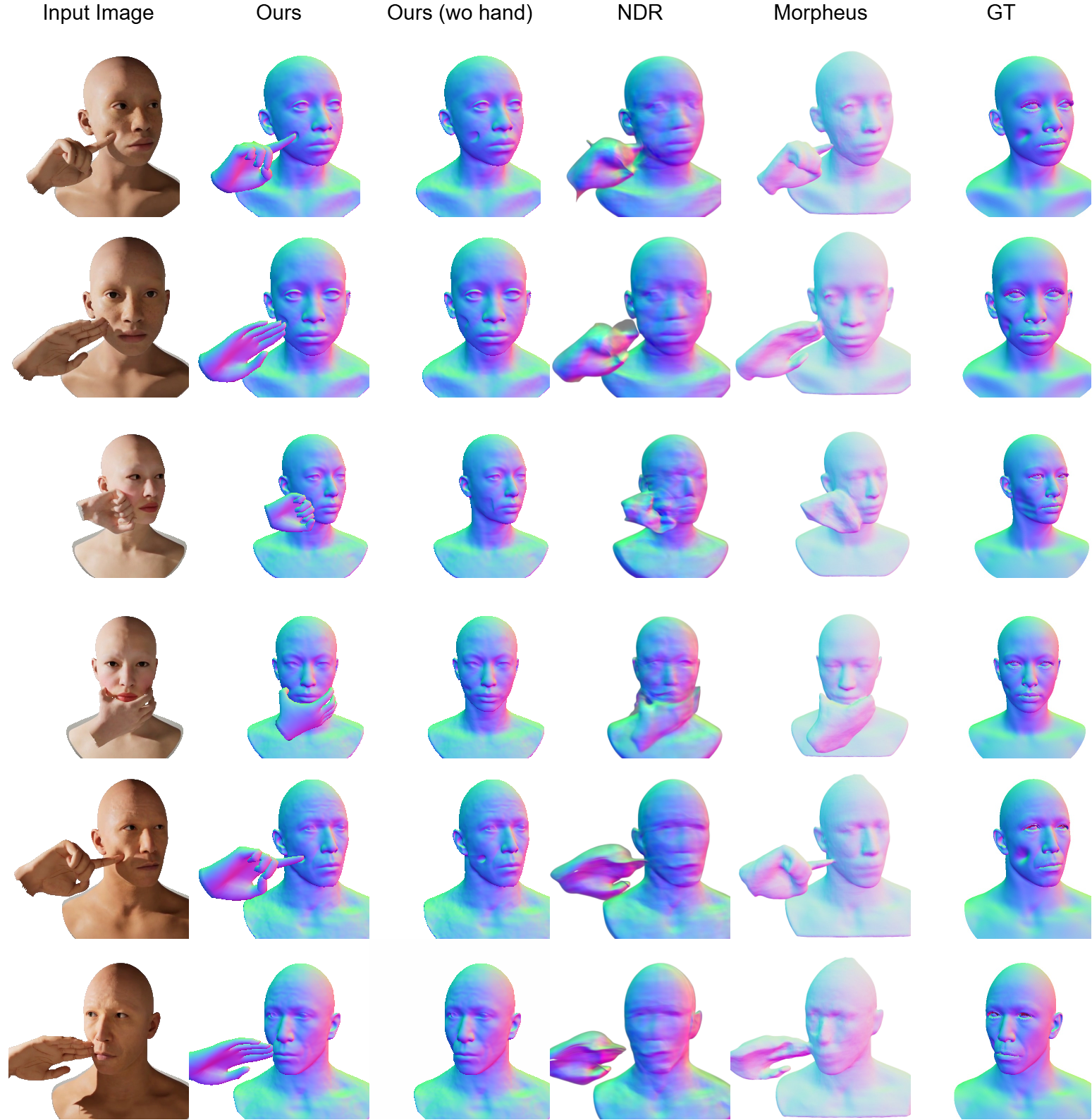}}
    \caption{
        \textbf{Qualitative Comparison with {SOTA}\xspace on Synthetic Videos} Reconstruction results from synthetic video sequences comparing our method with {SOTA}\xspace baselines. To ensure comparative fairness, we use the depth information in our method as NDR and Morpheus. Our approach achieves superior reconstruction of both hand geometry and facial features while maintaining physically consistent non-rigid deformations from hand-face interactions. The final column presents ground truth normal maps for reference, demonstrating our method's ability to recover intricate surface details.}
    \label{fig:qualitative_comparison_real_synthetic}
\end{figure*}%

\subsection{Comparison on Synthetic Dataset}



We conduct comprehensive qualitative and quantitative evaluations on synthetic videos. Unlike experiments with real-world captures, this synthetic dataset provides ground truth meshes corresponding to each frame, enabling precise metric-based evaluation of reconstruction accuracy. To ensure equitable benchmarking, we incorporate depth information when training our method with synthetic data.

As shown in Fig.~\ref{fig:qualitative_comparison_real_synthetic}, our method reconstructs detailed surface geometry. NDR fails to produce valid hand shapes and facial expressions, while Morpheus shows improved facial reconstruction but struggles with large hand motion variations and articulated hand shapes. Moreover, ours successfully models occluded regions, particularly hand-face contact zones, achieving deformation patterns (Column ~3) that closely match ground truth observations (last column).

For quantitative analysis, we extract meshes via marching cubes~\cite{we1987marching} from implicit surfaces and calculate metrics using ground truth meshes in our synthetic dataset. We evaluate reconstruction quality using four metrics: Chamfer Distance (CD) for global shape alignment, F-scores at 5~mm (F5) and 10~mm (F10) thresholds for local detail preservation, and Normal Consistency (NC) for surface orientation accuracy. As shown in Table~\ref{tab:quantitative_comparison}, our method outperforms baselines across all metrics. 
The highest Chamfer Distance and Normal Consistency indicate that our method captures the most accurate shape. F5 and F10 shows that our method is also the best in recovering shape details, which is consistent with the qualitative results.

\begin{table}
  \centering
  \begin{tabular}{l c c c c}
    \toprule
    Method & NC $\uparrow$ & CD $\downarrow$ & F5 $\uparrow$ & F10 $\uparrow$ \\
    \midrule
    NDR     & 52.35        & 19.14           & 0.54        & 1.93         \\
    Morpheus & 53.53       & 18.245          & 0.31        & 1.14         \\
    Ours    & \textbf{75.06} & \textbf{2.74}  & \textbf{10.22}        & \textbf{33.20}         \\
    \bottomrule
  \end{tabular}
  \caption{Quantitative comparison with NDR and Morpheus on our synthetic dataset.}
  \label{tab:quantitative_comparison}
\end{table}

\subsection{Ablation Studies}
We conducted comprehensive ablation studies to validate the contributions of individual components in our preprocessing and avatar reconstruction pipeline. 

\begin{figure}[t]
    \centerline{\includegraphics[width=1\linewidth]{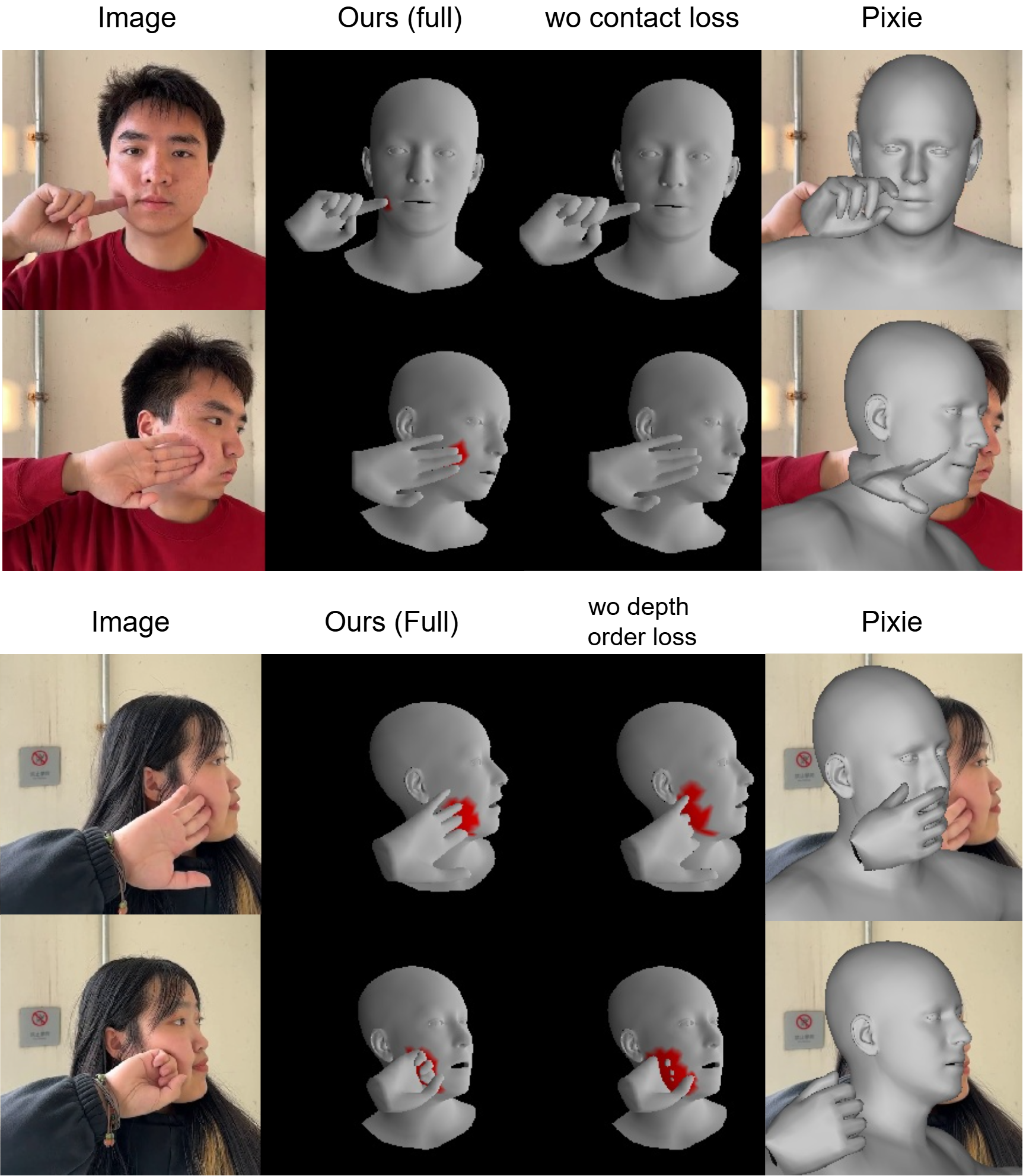}}
    \caption{
        \textbf{Ablation Study in the Preprocessing Stage} 
        The contact loss guides the hand mesh toward the surface of the face mesh to establish contact, as illustrated in Figure \ref{fig:ablation_preprocess}. Regions of contact between the meshes are visualized as red pixels on the face mesh. The depth order loss plays a critical role in ensuring plausible interactions by preventing excessive interpenetration of the meshes. We also compare our method with Pixie \cite{feng2021collaborative} in the last column.}
    \label{fig:ablation_preprocess}
\end{figure}

\noindent\textbf{Preprocessing Analysis}  
Our method extends beyond basic landmark alignment and temporal smoothing through two critical constraints: contact-aware alignment for hand-face proximity and depth-aware collision prevention for penetration avoidance.
Fig.~\ref{fig:ablation_preprocess} demonstrates these mechanisms through qualitative comparisons. The first two rows reveal how contact loss drives hand meshes toward facial surfaces, where the penetration between two meshes is represented using red pixels. The final two rows illustrate depth order loss's critical role in maintaining plausible spatial relationships, particularly for extreme head poses where two meshes tend to intersect too deep. 

The combination of these losses enables precise hand-face positioning that persists through dynamic interactions, providing reliable initialization for subsequent reconstruction stages. Qualitative comparisons against Pixie~\cite{feng2021collaborative} further confirm our method's superior capability in achieving accurate hand-face positions across diverse interaction scenarios.

\begin{figure}[t]
    \centerline{\includegraphics[width=1\linewidth]{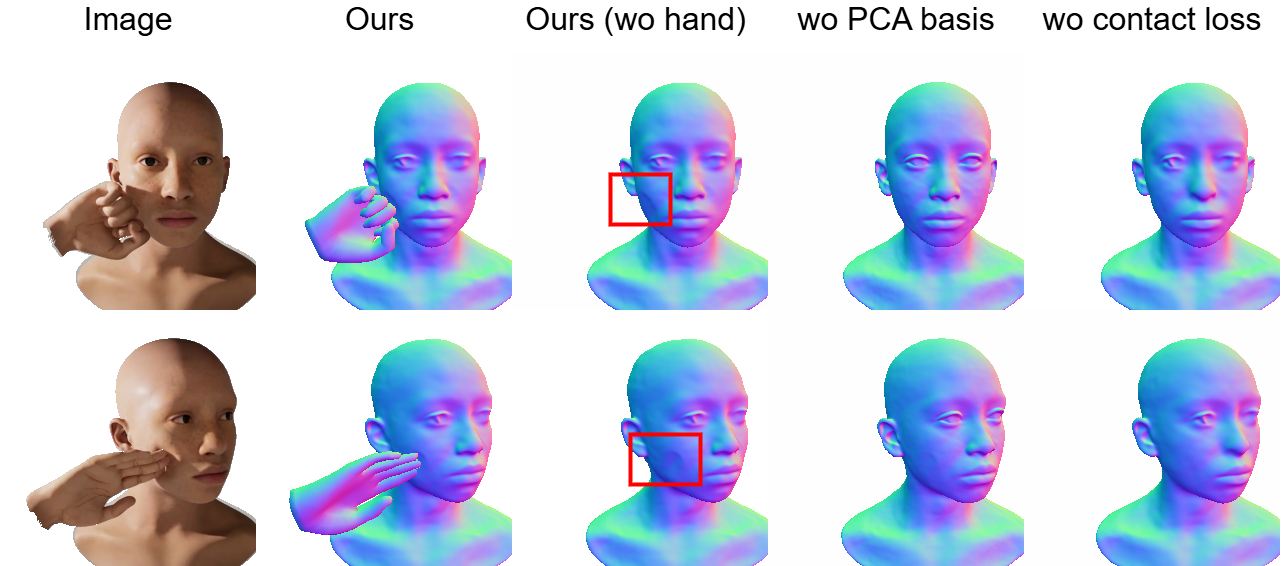}}
    \caption{
        \textbf{Ablation Study in the Reconstruction Stage} Learning from hand-face interaction PCA basis avoids predicting free spacial offsets, making the learning of non-rigid deformation much easier. Our contact loss is the key to achieving physically plausible facial non-rigid deformation caused by hand-face interaction.}
    \label{fig:ablation_reconstruction}
\end{figure}

\noindent\textbf{Reconstruction Analysis} During reconstruction, we leverage a PCA-based deformation prior derived from hand-face interaction data to learn non-rigid parameters and blendshapes. As demonstrated in Fig.~\ref{fig:ablation_reconstruction}, our PCA-driven approach (Column~3) achieves more natural facial deformations compared to direct spatial offset prediction (Column~4). The contact loss further plays a crucial role in regularizing physically plausible deformation (Column~5), completing our optimization framework.

\section{Conclusion}

We propose a method to reconstruct realistic head avatars with hand contact from monocular videos through two key components. First, we introduce contact loss and depth order loss during preprocessing to jointly align the hand and face mesh, establishing precise spatial relationships crucial for the subsequent stage. Second, we train a non-rigid deformation network that learns deformation parameters and blendshapes, supervised a PCA basis derived from hand-face interaction data, replacing direct spatial offset prediction with more efficient deformation learning. We also propose an additional contact loss to ensure physically plausible deformation results.

Our method successfully learns hand-face interactions from monocular input, but several areas remain for future exploration:
(1) The physics-inspired contact loss remains limited, as it cannot model effects such as skin pulling or friction.
(2) The material properties of skin, muscle, and fat are not explicitly modeled, leaving room for further exploration.
(3) The optimization process is slow, preventing real-time applications and highlighting the need for future research on accelerating interaction modeling.


{
    \small
    \bibliographystyle{ieeenat_fullname}
    \bibliography{main}

\begin{thebibliography}{30}
\providecommand{\natexlab}[1]{#1}
\providecommand{\url}[1]{\texttt{#1}}
\expandafter\ifx\csname urlstyle\endcsname\relax
  \providecommand{\doi}[1]{doi: #1}\else
  \providecommand{\doi}{doi: \begingroup \urlstyle{rm}\Url}\fi

\bibitem[Bharadwaj et~al.(2023)Bharadwaj, Zheng, Hilliges, Black, and Abrevaya]{bharadwaj2023flare}
Shrisha Bharadwaj, Yufeng Zheng, Otmar Hilliges, Michael~J. Black, and Victoria~Fernandez Abrevaya.
\newblock Flare: Fast learning of animatable and relightable mesh avatars.
\newblock \emph{ACM Transactions on Graphics}, 42:\penalty0 15, 2023.

\bibitem[Bozic et~al.(2020{\natexlab{a}})Bozic, Palafox, Zollh{\"o}fer, Dai, Thies, and Nie{\ss}ner]{bozic2020neural}
Aljaz Bozic, Pablo Palafox, Michael Zollh{\"o}fer, Angela Dai, Justus Thies, and Matthias Nie{\ss}ner.
\newblock Neural non-rigid tracking.
\newblock \emph{Advances in Neural Information Processing Systems}, 33:\penalty0 18727--18737, 2020{\natexlab{a}}.

\bibitem[Bozic et~al.(2020{\natexlab{b}})Bozic, Zollhofer, Theobalt, and Nie{\ss}ner]{bozic2020deepdeform}
Aljaz Bozic, Michael Zollhofer, Christian Theobalt, and Matthias Nie{\ss}ner.
\newblock Deepdeform: Learning non-rigid rgb-d reconstruction with semi-supervised data.
\newblock In \emph{Proceedings of the IEEE/CVF Conference on Computer Vision and Pattern Recognition}, pages 7002--7012, 2020{\natexlab{b}}.

\bibitem[Bulat and Tzimiropoulos(2017)]{bulat2017far}
Adrian Bulat and Georgios Tzimiropoulos.
\newblock How far are we from solving the 2d \& 3d face alignment problem?(and a dataset of 230,000 3d facial landmarks).
\newblock In \emph{Proceedings of the IEEE international conference on computer vision}, pages 1021--1030, 2017.

\bibitem[Cai et~al.(2022)Cai, Feng, Feng, Wang, and Zhang]{cai2022neural}
Hongrui Cai, Wanquan Feng, Xuetao Feng, Yan Wang, and Juyong Zhang.
\newblock Neural surface reconstruction of dynamic scenes with monocular rgb-d camera.
\newblock \emph{Advances in Neural Information Processing Systems}, 35:\penalty0 967--981, 2022.

\bibitem[Chen et~al.(2024)Chen, Wang, Li, Xiao, Zhang, Yao, and Liu]{chen2024monogaussianavatar}
Yufan Chen, Lizhen Wang, Qijing Li, Hongjiang Xiao, Shengping Zhang, Hongxun Yao, and Yebin Liu.
\newblock Monogaussianavatar: Monocular gaussian point-based head avatar.
\newblock In \emph{ACM SIGGRAPH 2024 Conference Papers}, pages 1--9, 2024.

\bibitem[Cheng et~al.(2024)Cheng, Oh, Price, Lee, and Schwing]{cheng2024putting}
Ho~Kei Cheng, Seoung~Wug Oh, Brian Price, Joon-Young Lee, and Alexander Schwing.
\newblock Putting the object back into video object segmentation.
\newblock In \emph{Proceedings of the IEEE/CVF Conference on Computer Vision and Pattern Recognition}, pages 3151--3161, 2024.

\bibitem[Duan et~al.(2023)Duan, Wang, Shi, Chen, and Cao]{Duan2023bakedavatar}
Hao-Bin Duan, Miao Wang, Jin-Chuan Shi, Xu-Chuan Chen, and Yan-Pei Cao.
\newblock Bakedavatar: Baking neural fields for real-time head avatar synthesis.
\newblock \emph{ACM Trans. Graph.}, 42\penalty0 (6), 2023.

\bibitem[Feng et~al.(2021{\natexlab{a}})Feng, Choutas, Bolkart, Tzionas, and Black]{feng2021collaborative}
Yao Feng, Vasileios Choutas, Timo Bolkart, Dimitrios Tzionas, and Michael~J Black.
\newblock Collaborative regression of expressive bodies using moderation.
\newblock In \emph{2021 International Conference on 3D Vision (3DV)}, pages 792--804. IEEE, 2021{\natexlab{a}}.

\bibitem[Feng et~al.(2021{\natexlab{b}})Feng, Feng, Black, and Bolkart]{feng2021learninganimatabledetailed3d}
Yao Feng, Haiwen Feng, Michael~J. Black, and Timo Bolkart.
\newblock Learning an animatable detailed 3d face model from in-the-wild images, 2021{\natexlab{b}}.

\bibitem[Izadi et~al.(2011)Izadi, Kim, Hilliges, Molyneaux, Newcombe, Kohli, Shotton, Hodges, Freeman, Davison, et~al.]{izadi2011kinectfusion}
Shahram Izadi, David Kim, Otmar Hilliges, David Molyneaux, Richard Newcombe, Pushmeet Kohli, Jamie Shotton, Steve Hodges, Dustin Freeman, Andrew Davison, et~al.
\newblock Kinectfusion: real-time 3d reconstruction and interaction using a moving depth camera.
\newblock In \emph{Proceedings of the 24th annual ACM symposium on User interface software and technology}, pages 559--568, 2011.

\bibitem[Juma~Rahman(2020)]{Rahman2020How}
Bapon~Fakhruddin Juma~Rahman, Jubayer~Mumin.
\newblock How frequently do we touch facial t-zone: A systematic review, 2020.
\newblock https://pmc.ncbi.nlm.nih.gov/articles/PMC7350942/.

\bibitem[Khirodkar et~al.(2024)Khirodkar, Bagautdinov, Martinez, Zhaoen, James, Selednik, Anderson, and Saito]{khirodkar2024sapiens}
Rawal Khirodkar, Timur Bagautdinov, Julieta Martinez, Su Zhaoen, Austin James, Peter Selednik, Stuart Anderson, and Shunsuke Saito.
\newblock Sapiens: Foundation for human vision models.
\newblock In \emph{European Conference on Computer Vision}, pages 206--228. Springer, 2024.

\bibitem[Li et~al.(2017)Li, Bolkart, Black, Li, and Romero]{li2017learning}
Tianye Li, Timo Bolkart, Michael~J Black, Hao Li, and Javier Romero.
\newblock Learning a model of facial shape and expression from 4d scans.
\newblock \emph{ACM Trans. Graph.}, 36\penalty0 (6):\penalty0 194--1, 2017.

\bibitem[Lin et~al.(2022)Lin, Zheng, Yong, and Xu]{lin2022occlusionfusion}
Wenbin Lin, Chengwei Zheng, Jun-Hai Yong, and Feng Xu.
\newblock Occlusionfusion: Occlusion-aware motion estimation for real-time dynamic 3d reconstruction.
\newblock In \emph{Proceedings of the IEEE/CVF Conference on Computer Vision and Pattern Recognition}, pages 1736--1745, 2022.

\bibitem[Newcombe et~al.(2015)Newcombe, Fox, and Seitz]{newcombe2015dynamicfusion}
Richard~A Newcombe, Dieter Fox, and Steven~M Seitz.
\newblock Dynamicfusion: Reconstruction and tracking of non-rigid scenes in real-time.
\newblock In \emph{Proceedings of the IEEE conference on computer vision and pattern recognition}, pages 343--352, 2015.

\bibitem[Pavlakos et~al.(2024)Pavlakos, Shan, Radosavovic, Kanazawa, Fouhey, and Malik]{pavlakos2024hamer}
Georgios Pavlakos, Dandan Shan, Ilija Radosavovic, Angjoo Kanazawa, David Fouhey, and Jitendra Malik.
\newblock Reconstructing hands in 3{D} with transformers.
\newblock In \emph{CVPR}, 2024.

\bibitem[Qian et~al.(2024)Qian, Kirschstein, Schoneveld, Davoli, Giebenhain, and Nie{\ss}ner]{qian2024gaussianavatars}
Shenhan Qian, Tobias Kirschstein, Liam Schoneveld, Davide Davoli, Simon Giebenhain, and Matthias Nie{\ss}ner.
\newblock Gaussianavatars: Photorealistic head avatars with rigged 3d gaussians.
\newblock In \emph{Proceedings of the IEEE/CVF Conference on Computer Vision and Pattern Recognition}, pages 20299--20309, 2024.

\bibitem[Romero et~al.(2022)Romero, Tzionas, and Black]{romero2022embodied}
Javier Romero, Dimitrios Tzionas, and Michael~J Black.
\newblock Embodied hands: Modeling and capturing hands and bodies together.
\newblock \emph{arXiv preprint arXiv:2201.02610}, 2022.

\bibitem[Shimada et~al.(2023)Shimada, Golyanik, P\'{e}rez, and Theobalt]{Shimada2023Decaf}
Soshi Shimada, Vladislav Golyanik, Patrick P\'{e}rez, and Christian Theobalt.
\newblock Decaf: Monocular deformation capture for face and hand interactions.
\newblock \emph{ACM Transactions on Graphics (TOG)}, 42\penalty0 (6), 2023.

\bibitem[Slavcheva et~al.(2017)Slavcheva, Baust, Cremers, and Ilic]{slavcheva2017killingfusion}
Miroslava Slavcheva, Maximilian Baust, Daniel Cremers, and Slobodan Ilic.
\newblock Killingfusion: Non-rigid 3d reconstruction without correspondences.
\newblock In \emph{Proceedings of the IEEE conference on computer vision and pattern recognition}, pages 1386--1395, 2017.

\bibitem[Slavcheva et~al.(2018)Slavcheva, Baust, and Ilic]{slavcheva2018sobolevfusion}
Miroslava Slavcheva, Maximilian Baust, and Slobodan Ilic.
\newblock Sobolevfusion: 3d reconstruction of scenes undergoing free non-rigid motion.
\newblock In \emph{Proceedings of the IEEE conference on computer vision and pattern recognition}, pages 2646--2655, 2018.

\bibitem[Wagner et~al.(2024)Wagner, Botsch, and Schwanecke]{wagner2024nephim}
Nicolas Wagner, Mario Botsch, and Ulrich Schwanecke.
\newblock Nephim: A neural physics-based head-hand interaction model, 2024.

\bibitem[Wang et~al.(2024)Wang, Wang, and Agapito]{wang2024morpheus}
Hengyi Wang, Jingwen Wang, and Lourdes Agapito.
\newblock Morpheus: Neural dynamic 360deg surface reconstruction from monocular rgb-d video.
\newblock In \emph{Proceedings of the IEEE/CVF Conference on Computer Vision and Pattern Recognition}, pages 20965--20976, 2024.

\bibitem[WE(1987)]{we1987marching}
LORENSEN WE.
\newblock Marching cubes: A high resolution 3d surface construction algorithm.
\newblock \emph{Computer graphics}, 21\penalty0 (1):\penalty0 7--12, 1987.

\bibitem[Wu et~al.(2024)Wu, Dou, Xu, Shimada, Wang, Yu, Liu, Lin, Cao, Komura, et~al.]{wu2024dice}
Qingxuan Wu, Zhiyang Dou, Sirui Xu, Soshi Shimada, Chen Wang, Zhengming Yu, Yuan Liu, Cheng Lin, Zeyu Cao, Taku Komura, et~al.
\newblock Dice: End-to-end deformation capture of hand-face interactions from a single image.
\newblock \emph{arXiv preprint arXiv:2406.17988}, 2024.

\bibitem[Xu et~al.(2024)Xu, Chen, Li, Zhang, Wang, Zheng, and Liu]{xu2023gaussianheadavatar}
Yuelang Xu, Benwang Chen, Zhe Li, Hongwen Zhang, Lizhen Wang, Zerong Zheng, and Yebin Liu.
\newblock Gaussian head avatar: Ultra high-fidelity head avatar via dynamic gaussians.
\newblock In \emph{Proceedings of the IEEE/CVF Conference on Computer Vision and Pattern Recognition (CVPR)}, 2024.

\bibitem[Yang et~al.(2024)Yang, Kang, Huang, Zhao, Xu, Feng, and Zhao]{yang2024depthv2}
Lihe Yang, Bingyi Kang, Zilong Huang, Zhen Zhao, Xiaogang Xu, Jiashi Feng, and Hengshuang Zhao.
\newblock Depth anything v2, 2024.

\bibitem[Yariv et~al.(2020)Yariv, Kasten, Moran, Galun, Atzmon, Ronen, and Lipman]{yariv2020multiview}
Lior Yariv, Yoni Kasten, Dror Moran, Meirav Galun, Matan Atzmon, Basri Ronen, and Yaron Lipman.
\newblock Multiview neural surface reconstruction by disentangling geometry and appearance.
\newblock \emph{Advances in Neural Information Processing Systems}, 33:\penalty0 2492--2502, 2020.

\bibitem[Zheng et~al.(2023)Zheng, Yifan, Wetzstein, Black, and Hilliges]{Zheng2023pointavatar}
Yufeng Zheng, Wang Yifan, Gordon Wetzstein, Michael~J. Black, and Otmar Hilliges.
\newblock Pointavatar: Deformable point-based head avatars from videos.
\newblock In \emph{Proceedings of the IEEE/CVF Conference on Computer Vision and Pattern Recognition (CVPR)}, 2023.

\end{thebibliography}
}


\end{document}